%% file: arxiv.tex
\ifdefined\XeTeXrevision\else
  \ifdefined\pdfoutput\pdfoutput=1\fi
\fi
\documentclass[11pt]{article}

\usepackage{tdgrokking_arxiv}

\usepackage[utf8]{inputenc}
\usepackage[T1]{fontenc}
\usepackage[hidelinks]{hyperref}
\usepackage{url}
\usepackage{booktabs}
\usepackage{array}
\usepackage{amsfonts}
\usepackage{amsmath}
\usepackage{amssymb}
\usepackage{graphicx}
\usepackage[font=small,labelfont=bf,labelsep=period]{caption}
\usepackage{nicefrac}
\usepackage{microtype}
\usepackage{xcolor}
\usepackage{listings}
\usepackage{wrapfig}

\newcommand{\method}{TD-Grokking}
\lstdefinestyle{prompttemplate}{
  basicstyle=\ttfamily\scriptsize,
  breaklines=true,
  breakatwhitespace=false,
  columns=fullflexible,
  keepspaces=true,
  showstringspaces=false,
  frame=single,
  rulecolor=\color{black!35},
  backgroundcolor=\color{black!3},
  xleftmargin=0.8em,
  xrightmargin=0.8em,
  aboveskip=0.75em,
  belowskip=0.75em
}

\title{TD-Grokking: Learning from Zero-Reward Problems by Training-Time Decomposition}
\author{
Ningyuan Xi\textsuperscript{1,2} \quad
Hao Xu\textsuperscript{3} \quad
Hongsheng Xin\textsuperscript{3} \quad
Ning Miao\textsuperscript{1,2,\ensuremath{\dagger}}
}
\date{}

\begin{document}

\maketitle
\footnotetext[1]{Department of Data Science, City University of Hong Kong.}
\footnotetext[2]{Hong Kong Institute of AI for Science, City University of Hong Kong.}
\footnotetext[3]{Li Auto Inc.}
\begingroup
\renewcommand{\thefootnote}{\ensuremath{\dagger}}
\footnotetext{Corresponding author: Ning Miao (\texttt{ningmiao@cityu.edu.hk}).}
\endgroup

\begin{abstract}
\input{sections/abstract}
\end{abstract}

\section{Introduction}
\input{sections/introduction}

\section{Related Work}
\input{sections/related_work}

\section{Method}
\input{sections/method}

\section{Experiments}
\input{sections/experiments}

\section{Analyzing the Learning Mechanism of {\method}}
\input{sections/analysis}

\section{Conclusion, Limitations, and Future Work}
\input{sections/conclusion}
\input{sections/limitations}

\bibliographystyle{plainnat}
\bibliography{refs}

\appendix
\input{sections/appendix}

\end{document}

%% file: sections/abstract.tex
Large language models (LLMs) have made remarkable progress in reasoning tasks, largely driven by post-training paradigms, especially reinforcement learning with verifiable rewards (RLVR). However, a critical bottleneck persists: RLVR fails on highly challenging zero-reward problems, where all sampled reasoning trajectories yield uniformly failed outcomes, providing no optimization signal to drive model improvement. 
Prior efforts to address this limitation, such as dense process supervision, partial reward assignment, or prefix-guided exploration, suffer from inherent task constraints or do not fully equip the policy model with the capabilities necessary to solve the original intractable problems.
To address this, we propose TD-Grokking, a training-time decomposition framework for zero-reward problems. It recursively decomposes intractable root problems into self-contained, verifiable subproblems, forming hierarchical trees where solvable leaves provide non-zero rewards. 
Evaluations on mathematical and medical tasks show that TD-Grokking outperforms vanilla GRPO as well as all baseline approaches.
Together with detailed analysis, these results confirm that training-time decomposition effectively converts zero-reward examples into usable training signals, enabling consistent performance gains. Our code and datasets are available at
\url{https://anonymous.4open.science/r/TD-Grokking-6567/}.

%% file: sections/introduction.tex
Large language models~(LLMs) have achieved remarkable progress in mathematical and algorithmic reasoning, with performance continuing to advance rapidly following elaborate post-training paradigms \citep{shao2024deepseekmath, guo2025deepseekr1, yang2025qwen3}.
As an essential component of post-training, reinforcement learning with verifiable rewards~(RLVR) unlocks the inherent reasoning capabilities of LLMs by contrasting successful and unsuccessful reasoning trajectories \citep{shao2024deepseekmath, guo2025deepseekr1}.
This naturally raises a fundamental research question: how can LLMs acquire the ability to tackle highly challenging problems where no successful trials are available?
Learning from such zero-reward problems poses a critical bottleneck: training on these questions offers no optimization signal, as all sampled rollouts yield uniformly failed outcomes. The resulting constant zero reward signal stagnates model optimization and renders standard RLVR ineffective.

Prior research has made considerable efforts to alleviate the unlearnability of zero-reward problems.
For example, \citet{lightman2023lets} demonstrate that dense process supervision, such as process reward models (PRMs), provides scores for intermediate steps in a generated reasoning trajectory, thereby refining credit assignment beyond final-answer supervision.
In code generation tasks, \citet{sun2025rlgrok} introduce the notion of partial correctness by giving partial credits to code generations that pass a subset of test cases. Their observations reveal a grokking-like phenomenon: full-pass reward starts to increase after roughly 450 RL training steps, mirroring the \textbf{grokking} phenomenon observed in supervised learning.
Despite their empirical effectiveness, process or partial reward strategies suffer from inherent limitations and task constraints. In mathematical reasoning, training accurate PRMs on frontier problems remains difficult, and evaluating partial correctness for a solution is often infeasible.

A separate line of research facilitates exploration on hard problems by conditioning the model generation on partial solution trajectories or privileged hints \citep{li2025questa, zhang2025scafgrpo, chen2025nurl, liao2026sage, xia2026hill}.
Such prefix-guided exploration allows the policy model to focus on the latter part of the solution, which increases the chance of obtaining non-zero rewards.
While effective as an exploration aid, this approach shortens the horizon of the current rollout rather than endowing the model with all core capabilities required to solve the original problem.
In other words, the model learns to continue reasoning from an intermediate state with partially solved trajectories, but the upstream reasoning required to reach those intermediate states remains unmodeled.

\begin{figure*}[t]
    \centering
    \includegraphics[width=\textwidth]{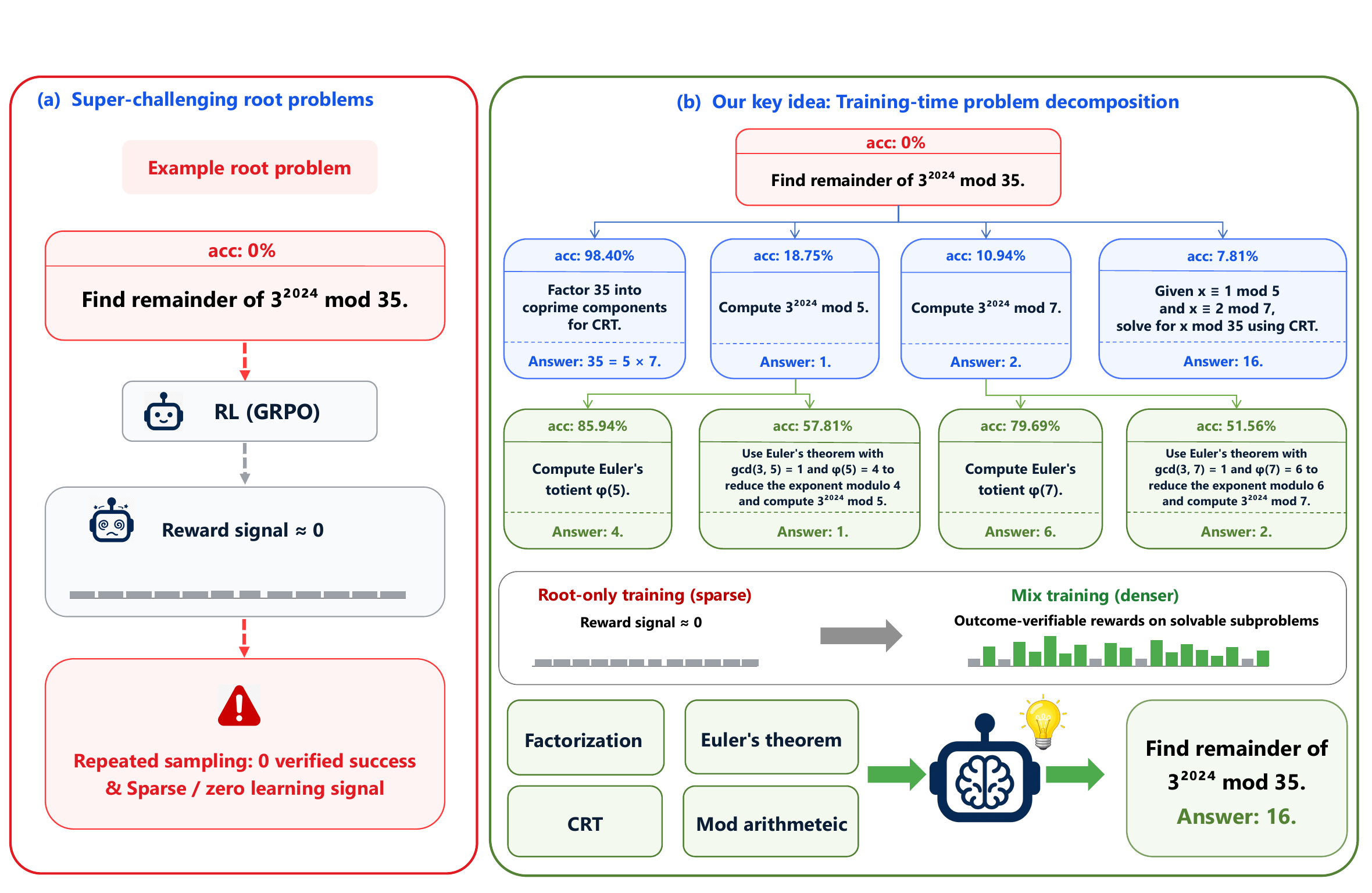}
    \caption{
    Overview of {\method}.
    (a) Regular GRPO stagnates on challenging zero-reward problems. (b) Through training-time decomposition, {\method} obtains dense training signals.
    }
    \label{fig:overview}
\end{figure*}

In this work, we propose a training-time decomposition framework, \textbf{TD-Grokking}, which is tailored specifically for effective learning from challenging zero-reward problems.
For each intractable zero-reward root problem, we use a decomposition generator to construct self-contained, verifiable subproblems that encapsulate the sub-capabilities essential for solving the original problem.
If a generated subproblem remains beyond the capability of the current policy model, we recursively apply the decomposition procedure.
This pipeline transforms intractable zero-reward problems into hierarchical decomposition trees, where solvable leaf subproblems provide non-zero outcome rewards under standard final-answer verification.
RL training initiates from these reward-enriched leaf nodes, and progressive optimization propagates upward to enhance performance on parent nodes, including root nodes, once child subproblems are reliably solved. Figure~\ref{fig:overview} illustrates our decomposition-based training pipeline.

Empirically, we evaluate {\method} across mathematical and medical domains, both of which contain highly challenging datasets with zero-reward problems.
On mathematical benchmarks, {\method} improves the accuracy on AIME 24 and 25 by approximately 4\% over vanilla GRPO~\citep{shao2024deepseekmath}, outperforming all baseline approaches.
On medical tasks, {\method} surpasses vanilla GRPO by up to 6.2\%.
These results demonstrate that decomposition acts as an effective training-time mechanism to convert zero-reward examples into usable learning signals, yielding consistent performance gains on challenging benchmarks.

%% file: sections/related_work.tex


We now introduce previous efforts to help LLMs tackle hard problems. We start with inference-time boosting of LLMs, and then discuss two major approaches to learning to solve hard problems.

\paragraph{Inference-time problem decomposition.}
Decomposition has also been widely used at inference time.
Chain-of-thought aggregates intermediate reasoning paths \citep{wei2022chain}.
Least-to-most prompting, decomposed prompting, Self-Ask, and Plan-and-Solve further structure inference through subproblems, plans, modular prompts, or tool-mediated steps \citep{zhou2023leasttomost, khot2023decomposed, press2023selfask, wang2023plan}.
Tree-of-thought methods extend this idea by searching over multiple intermediate reasoning states with lookahead, backtracking, or self-evaluation \citep{yao2023tree}.

These methods use decomposition as an inference-time scaffold for a fixed model.
They can improve how the model organizes reasoning, but they do not directly train required but missing reasoning skills.
They are therefore most effective when the required subskills are already within the model's reachable repertoire.
Our setting is different: the initial model receives nearly zero verified reward on the root problems.
We therefore use decomposition as a training-time mechanism, where each subproblem becomes an independent RL instance rather than an intermediate prompt within a single inference attempt.

\paragraph{Process-level and partial feedback.}
When a question is so hard that the LLM cannot reach a correct answer after multiple trials, the reward will be zero for all generations, which makes learning impossible.
The first approach to mitigate this is to give rewards to partially correct solutions.
For example, process supervision trains reward models to score intermediate reasoning steps instead of only final answers, and has been shown to improve mathematical reasoning and verification \citep{lightman2023lets}.
Follow-up work reduces the annotation cost by automatically constructing step-level labels or collecting process reward data through search, as in Math-Shepherd and OmegaPRM \citep{wang2024mathshepherd, luo2024omegaprm}.
However, it is very costly to train a separate process reward model, and the accuracy and generalization of existing PRMs make them unsuitable for the RL training of SOTA models \citep{zheng2024processbench}.
In coding tasks, special partial-correctness signals can also be obtained from subsets of test cases, allowing RL to reward programs before full correctness is achieved \citep{sun2025rlgrok}.



\paragraph{Training-time hints and scaffolded exploration.}
A related line of work improves exploration by making difficult problems easier during training.
QuestA augments hard questions with partial solution sketches \citep{li2025questa}; Scaf-GRPO injects hierarchical hints when GRPO encounters all-fail rollout groups \citep{zhang2025scafgrpo}; NuRL and self-hinting methods use generated cues to move hard prompts into a learnable region \citep{chen2025nurl, liao2026sage}; and HiLL studies whether hinted success transfers to the no-hint setting \citep{xia2026hill}.

These methods are effective exploration aids, but a hint-conditioned trajectory is not the same as solving the original root problem.
The model may learn to continue from a useful hint without learning to produce the hidden reasoning state itself.
In contrast, our method uses decomposition not merely to shorten the rollout horizon, but to convert hidden reasoning requirements into subproblems with independent verifiable rewards.

%% file: sections/method.tex
The core objective of {\method} is to extract usable training signal from challenging reasoning problems that produce no useful outcome reward under standard direct RLVR. 
Given a difficult reasoning problem, {\method} does not modify the final-answer verifier, introduce learned process reward models, or make root rollouts easier by supplying privileged hints.
Instead, it reorganizes the fundamental training unit: zero-reward problems are expanded into smaller, self-contained, fully verifiable subproblems that capture the reasoning demands required to solve the original root problem. 
Reinforcement learning is then performed on these subproblems with standard outcome-based rewards, enabling the model to escape the zero-reward regime and recover performance on the original root problems.

\subsection{Problem Setting}

Let $x$ denote a root reasoning problem with ground-truth verifiable answer $y$.
A policy model $\pi_\theta$ generates a solution trajectory
$o \sim \pi_\theta(\cdot \mid x)$, from which an answer $\hat{y}(o)$ is extracted.
The standard outcome reward is
\[
R(x,o)=\mathbf{1}\left[
\operatorname{Verify}\bigl(\hat{y}(o),y\bigr)=1
\right].
\]

For any verifiable problem $q$ with target answer $a$, we define its empirical verified accuracy under policy $\pi$ and sampling budget $K$ as
\[
\operatorname{Acc}^{\pi}_{K}(q)=
\frac{1}{K}\sum_{k=1}^{K}
\mathbf{1}\left[
\operatorname{Verify}\bigl(\hat{a}(o_k),a\bigr)=1
\right],
\qquad
o_k \sim \pi(\cdot \mid q).
\]
The problem $q$ may be either an original root problem or a decomposed subproblem.
For an initial policy $\pi_{\theta_0}$, a root problem $x$ is called \emph{zero-reward} under budget $K$ if
\[
\operatorname{Acc}^{\pi_{\theta_0}}_{K}(x)=0.
\]
This definition is policy- and budget-dependent: the problem is not assumed to be intrinsically unsolvable, but it is uninformative for direct outcome-only RL because all sampled rollouts receive zero reward.

\subsection{Constructing Verifiable Subproblems}

For each hard root problem $x$, {\method} generates a set of labeled candidate instances
\[
\mathcal{D}(x)=\{(s_{x,1},a_{x,1}),\ldots,(s_{x,m_x},a_{x,m_x})\},
\]
where $s_{x,j}$ denotes a candidate subproblem and $a_{x,j}$ is its target answer used for verification. We use the term \emph{subproblem} to refer to the question $s_{x,j}$ itself, while the pair $(s_{x,j},a_{x,j})$ denotes the corresponding verifiable training instance.

A candidate subproblem $s_{x,j}$ is retained only if it satisfies three conditions. First, it is \emph{root-conditioned}: it corresponds to a local reasoning requirement used in solving the parent root, rather than to a generic skill label. Second, it is \emph{self-contained}: all assumptions needed to solve it are stated in the subproblem itself, with no dependence on the root solution or on other subproblem answers. Third, it is \emph{verifiable}: its final answer has a well-defined target $a_{x,j}$ and can be checked by the same outcome-style verifier used for RLVR, after the usual answer extraction and normalization.

The decomposition pipeline is implemented as a sequence of structured calls to the decomposition generator, followed by validation. Exact prompt templates, parsing rules, and generation hyperparameters are reported in the appendix. The main method has six stages.

\paragraph{1. Hard-root selection.}
We first identify roots on which the starting policy receives zero verified reward under repeated sampling. This focuses decomposition on examples that direct outcome-only RL is least able to exploit. The selection criterion is tied to the base policy, verifier, and sampling budget; a root can leave the zero-reward set after training.

\paragraph{2. Guide preparation.}
For each selected root, {\method} obtains a decomposition guide as described above. When a dataset-provided solution is available, the guide is taken from the source data after answer-consistency checking. When it is unavailable, the decomposition generator first produces a solution sketch whose final answer must verify against the known target answer. This step makes the subsequent decomposition solution-guided rather than purely associative: subproblems are extracted from a concrete reasoning path instead of being sampled as generic skills that merely look related to the root.

\paragraph{3. Solution-guided segmentation.}
Given $(x,y,g)$, the generator segments the guide into a short ordered list of local reasoning requirements. A segment is not a training label and is not exposed to the student. Its purpose is to locate the operations that make the root difficult as an end-to-end problem: for example, identifying a hidden constraint, deriving a recurrence, simplifying a symbolic expression, resolving a case split, or computing a key intermediate value.

\paragraph{4. Requirement extraction.}
For each segment, the generator extracts the minimal information needed to state the corresponding local requirement as a separate task. This includes relevant conditions from the original problem statement and, when necessary, intermediate facts from the guide. The extraction step separates \emph{given conditions}, which should appear in the subproblem statement, from \emph{target conclusions}, which should remain for the policy to solve. This distinction is what prevents decomposition from becoming either underspecified local hints or a copied solution trace.

\paragraph{5. Self-contained rewriting.}
The extracted requirement is rewritten into a problem--answer pair $(s_{x,j},a_{x,j})$. The rewritten problem must contain all assumptions needed for solving it, avoid references to neighboring subproblems, and end with an answer that can be automatically checked. This step turns a latent operation in the root solution into an independent RL instance.

\paragraph{6. Structured validation.}
The raw decomposition is not used directly. {\method} treats validation as part of the method, because the useful signal comes from retaining only those local tasks that are simultaneously root-derived, self-contained, and verifiable. The validator checks candidates at multiple levels.

At the \emph{format level}, each candidate must expose a parseable problem, answer, and optional solution field; the answer must be non-empty and concise rather than a paragraph of explanation. At the \emph{dependency level}, the problem must be self-contained, must include all necessary assumptions, and must avoid references to earlier subproblems or to the guide. At the \emph{verifier level}, the answer must be compatible with the same answer-extraction and normalization interface used for the root. At the \emph{structure level}, the retained decomposition should cover meaningful local requirements from the guide without excessive duplication, over-fragmentation, direct root copying, or answer-revealing shortcuts.

Only candidates that pass validation are used for training. If too few candidates pass, or if the retained candidates do not cover the major reasoning requirements in the guide, the root is regenerated with stricter instructions. This retry loop is a quality-control step: it improves the reliability of the decomposed training pool while preserving the invariant that every training node is a self-contained, verifiable task.

\subsection{Difficulty Calibration and Recursive Expansion}

After validation, {\method} calibrates the difficulty of each retained node against the current policy. For a retained subproblem $s_{x,j}$, we estimate $\operatorname{Acc}^{\pi}_{K_c}(s_{x,j})$ using the same outcome verifier as above. A subproblem with at least one verified rollout under the calibration budget, i.e., $\operatorname{Acc}^{\pi}_{K_c}(s_{x,j})>0$, is treated as an \emph{active trainable leaf}: it is not already solved by construction, but it can provide nonzero reward events for RL. A retained subproblem with $\operatorname{Acc}^{\pi}_{K_c}(s_{x,j})=0$ is an unresolved node for the current policy. Such a node can be decomposed again using the same guide-preparation, segmentation, rewriting, and validation procedure, or deferred until the policy improves.

Thus, {\method} naturally defines a decomposition tree. The original hard problem is the root, unresolved subproblems become internal nodes, and calibrated trainable subproblems become leaves. Although the procedure allows recursive expansion, in practice we found that most first-level retained subproblems already produced verified rollouts under the starting policy.

\subsection{Training on the Decomposition Tree}

Given a decomposition tree, {\method} trains the policy bottom-up. At a given training stage, the active set consists of retained nodes that are verifiable and appropriate for the current policy's difficulty level. Training first emphasizes active leaves, because they provide outcome rewards while remaining tied to the original hard root. After training, the policy is re-evaluated on parent nodes and on the root. If a parent begins to produce verified rollouts, it can be promoted into the active set so that the model practices recomposing the learned local requirements into a larger reasoning unit. If a parent remains zero-reward, the same decomposition operation can be applied again or the node can remain deferred. This yields a curriculum in which the model first learns local requirements and then recomposes them into progressively larger reasoning units.

All nodes use the same final-answer RLVR interface. For a node $z$, whether it is a root or a subproblem, the rollout receives reward only if its extracted final answer verifies against the node's target $a_z$. {\method} therefore does not require process labels, partial-credit rewards, or task-specific reward shaping. The only change is which verifiable items are presented to RL and in what order.


Because calibration showed that first-level subproblems were already trainable, our controlled experiments stop expansion after the first level to isolate the effect of root-derived subproblem training. We report three static training views over the same source roots. \textbf{Root-only} is exactly vanilla RLVR on the original root problems, using the same outcome verifier, reward definition, prompting format, and optimizer, but without any decomposition-derived items. \textbf{Sub-only} trains only on the retained subproblems and serves as the diagnostic for local-to-global recomposition, because any root-level recovery cannot come from direct root rewards. \textbf{Mixed} trains on the union of roots and their subproblems, preserving pressure on the final task while adding local outcome-reward signal from easier, root-derived items. In addition, we evaluate a \textbf{Dynamic} variant that maintains an accuracy-based active set: subproblems are introduced when their parent remains uninformative and are retired once they become reliably solved, reducing training cost while following the same decomposition tree. These views are experimental controls over the same decomposition tree, rather than separate reward objectives.


%% file: sections/experiments.tex
We apply {\method} to improve training on zero-reward problems in two distinct domains.
On competition-level math, we test the main hypothesis: whether training-time decomposition can turn otherwise uninformative roots into useful RL training signals.
We then turn to the medical domain to test whether {\method} remains useful outside mathematical reasoning.
Across all experiments, decomposition is performed only during training-data construction; evaluation is performed on the original benchmark questions without inference-time decomposition.

\subsection{Mathematical Reasoning}

\paragraph{Setup.}
All mathematical {\method} variants start from Qwen3-1.7B and use GRPO with a final-answer verifiable reward \citep{yang2025qwen3, shao2024deepseekmath}. The training pool is constructed from $3{,}788$ DeepScaleR-hard questions, and the corpus is described in Appendix~\ref{app:corpus}.
We compare our model with multiple baselines on benchmarks including AIME 2024/2025, AMC23, OlympiadBench, MATH500, and GPQA-Diamond \citep{aime24, aime25, amc23, he2024olympiadbench, hendrycks2021math, rein2024gpqa}.
We report four variants of {\method}: \textbf{Root-only}, \textbf{Sub-only}, \textbf{Mixed}, and \textbf{Dynamic}. 
Root-only, reported as \textbf{Vanilla} in the following results, trains on the original zero-reward roots and is identical to standard RLVR on the root problems. 
Sub-only trains only on root-derived subproblems and serves as the diagnostic for local-to-global recomposition, because the checkpoint never receives root rewards during training. 
Mixed trains on the union of roots and subproblems, preserving pressure on the final root task while adding trainable local leaves.

To verify that the performance gain from {\method} is not simply an artifact of adding more training examples compared with GRPO on original questions, we add a comparative experiment named GRPO-simple.
Specifically, we train GRPO-simple by adding a comparable-scale, comparable-difficulty MWP-RLVR set whose examples have no decomposition relationship to the selected zero-reward roots.
The set contains $11{,}896$ examples from \textsc{Dolphin18K-clean-single-numeric}, $2{,}373$ from \textsc{MAWPS}, and $1{,}218$ from \textsc{ASDiv-A}, for a total of $15{,}487$ training rows. 
These source corpora are drawn from established math word-problem datasets \citep{huang2016dolphin18k, koncelkedziorski2016mawps, miao2020asdiv}.

We further compare {\method} with an SFT model tuned on the same reference solutions used for subproblem decomposition.

\begin{figure*}[t]
\centering
\begin{minipage}[t]{0.49\textwidth}
\centering
\includegraphics[width=\linewidth]{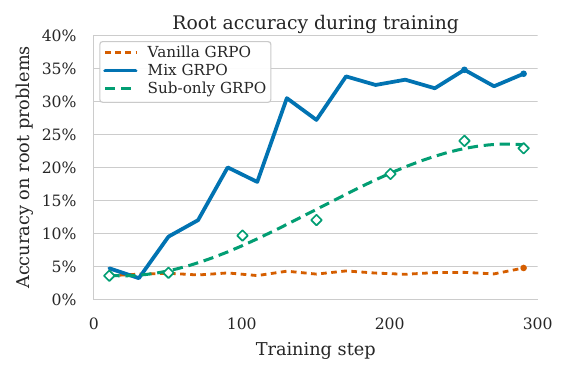}
\end{minipage}\hfill
\begin{minipage}[t]{0.49\textwidth}
\centering
\includegraphics[width=\linewidth]{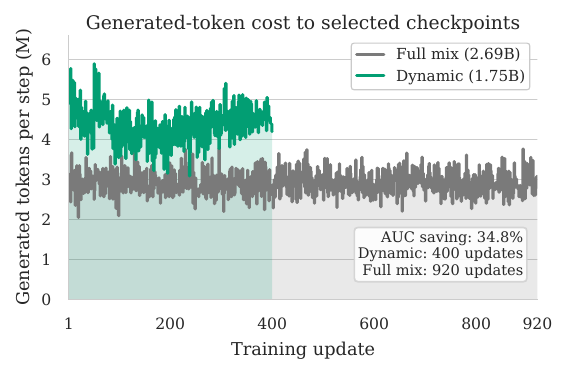}
\end{minipage}
\caption{
Training-time root accuracy and generated-token cost. Left: accuracy on original root prompts over 20-step windows. Vanilla GRPO denotes direct GRPO on the original root problems without subproblem augmentation; Mixed GRPO keeps root prompts in the training mixture, and Sub-only GRPO is shown as the subproblem-only comparison. Right: cumulative generated response-token AUC from run start to the selected checkpoints, showing that dynamic mixing uses fewer generated tokens than the full mixed run in this accounting.
}
\label{fig:mix_root_training_curve}
\label{fig:dynamic_auc}
\end{figure*}

\paragraph{Main results.}
Figure~\ref{fig:mix_root_training_curve} summarizes both the root-level training dynamics and the generated-token cost of dynamic mixing. The training-accuracy panel is evaluated only on original root questions.
We can see that training only on very hard original questions leads to little performance gain.
In comparison, RL training on their subproblems activates learning and equips the LLM with abilities that transfer back to original problems.
By mixing original questions with subproblems, the LLM can practice local skills while remaining aligned with the target root-question distribution, leading to stronger root-level performance.

Table~\ref{tab:main_math_results} summarizes the performance of {\method} and baseline approaches on mainstream mathematical benchmarks. 
Among the three primary static TD-Grokking variants, mixed training achieves the strongest performance on all reported mathematical benchmarks except MATH500. 
For example, on AIME 2024/2025, mixed {\method} improves avg@8 accuracy by 6.25 and 3.75 percentage points compared with vanilla GRPO training on original problems. 
These results show that subproblems provide denser reward signals, which makes training on originally zero-reward roots more effective.
Original problems are also important to keep the policy aligned with the target question distribution.

\begin{table*}[t]
\caption{
Experimental results on mathematical and science reasoning benchmarks. All entries are percentages. For AIME, avg@8 is mean correctness over eight sampled rollouts per problem, and pass@8 counts whether at least one sampled rollout solves the problem. Vanilla, sub-only, and mixed are the primary static TD-Grokking views. Bold marks the best completed result in each column.
}
\centering
\small
\setlength{\tabcolsep}{3.2pt}
\resizebox{\textwidth}{!}
{%
\begin{tabular}{lcccccccc}
\toprule
\textbf{Model}
& \multicolumn{2}{c}{\textbf{AIME 2024}}
& \multicolumn{2}{c}{\textbf{AIME 2025}}
& \textbf{Olympiad}
& \textbf{MATH500}
& \textbf{AMC23}
& \textbf{GPQA-Diamond} \\
\cmidrule(lr){2-3}
\cmidrule(lr){4-5}
\cmidrule(lr){6-9}
& \textbf{avg@8}
& \textbf{pass@8}
& \textbf{avg@8}
& \textbf{pass@8}
& \multicolumn{4}{c}{\textbf{pass@1}} \\
\midrule
Base
& 42.08 & 70.00 & 35.83 & 53.33
& 56.53 & 88.80 & 73.75 & 36.36
\\
GRPO
& & & &
&  &  &  &
\\
\quad Vanilla
& 42.08 & 76.67 & 36.67 & 56.67
& 56.23 & 89.80 & 77.50 & 34.34
\\
\quad Ablation
& 46.67 & 73.33 & 34.17 & 50.00
& 57.42 & 89.60 & 75.00 & 36.87
\\
SFT
& 43.75 & 73.33 & 37.50 & 50.00
& 57.57 & 88.60 & 75.00 & 33.84
\\
QuestA
& 47.50 & 76.67 & 37.50 & 60.00
& 58.31 & 89.00 & 77.50 & 38.89
\\
Scaf-GRPO
& 42.08 & 73.33 & 34.58 & \textbf{63.33}
& 57.86 & \textbf{90.40} & 76.25 & 38.38
\\

\textbf{\method}
& & & &
&  &  &  &
\\
\quad Sub-only
& 40.42 & 73.33 & 34.58 & 50.00
& 58.01 & 87.60 & 79.58 & 34.34
\\
\quad Mixed
& \textbf{48.33} & \textbf{80.00} & \textbf{40.42} & 60.00
& \textbf{59.64} & 90.20 & \textbf{83.33} & \textbf{39.90}
\\
\quad Dynamic
& 47.91 & \textbf{80.00} & 39.58 & \textbf{63.33}
& 58.01 & 90.00 & 80.83 & 38.89
\\
\bottomrule
\end{tabular}%
}
\label{tab:main_math_results}
\end{table*}

Through dynamic problem selection, dynamic mixing can achieve performance comparable to fully mixed training with lower generated-token cost. In the checkpoint-cumulative accounting in Figure~\ref{fig:dynamic_auc}, dynamic mixing saves about 35\% of generated response tokens compared with fully mixed training.
Further comparisons with GRPO-simple and SFT show that the improvement of {\method} is not merely a result of more training data or the use of reference CoT solutions. The training-time decomposition strategy is the most critical factor in the improved reasoning performance.

\subsection{Medical-Domain Instantiation}
In this section, we investigate whether {\method} remains useful outside mathematics.
To this end, we choose the medical domain, which requires probabilistic and case-based reasoning and is very different from mathematical reasoning.
We compare TD-Grokking with all baseline approaches on the challenging MedBullet dataset \citep{chen2025benchmarking}, following the same setup as in the mathematical-reasoning experiments.
We evaluate on MedQA, MedMCQA, PubMedQA, and MMLU medical subsets \citep{jin2021medqa, pal2022medmcqa, jin2019pubmedqa, hendrycks2021mmlu}.


\begin{wraptable}{r}{0.48\textwidth}
\vspace{-0.8em}
\centering
\small
\setlength{\tabcolsep}{3.2pt}
\caption{
Medical-domain results on MedBullet. All entries are percentages. 
\textsc{MMLU Medical} is the macro-average over five medical-related MMLU domains.
}
\resizebox{\linewidth}{!}{%
\begin{tabular}{lcccc}
\toprule
\textbf{Benchmark}
& \textbf{Base}
& \textbf{Vanilla}
& \textbf{Sub-only}
& \textbf{Mixed} \\
\midrule
\textsc{MedQA} & 27.10 & 29.14 & 27.10 & \textbf{31.26} \\
\textsc{MedMCQA} & 39.59 & 39.59 & 39.59 & \textbf{41.67} \\
\textsc{PubMedQA} & 54.20 & 52.10 & 52.10 & \textbf{58.30} \\
\textsc{MMLU Medical} & 52.60 & 53.44 & 52.60 & \textbf{55.91} \\
\midrule
Average & 43.37 & 43.57 & 42.85 & \textbf{46.78} \\
\bottomrule
\end{tabular}%
}

\label{tab:medical_domain}
\vspace{-1.0em}
\end{wraptable}

Table~\ref{tab:medical_domain} shows the same qualitative pattern as the mathematical experiments. 
Mixed training outperforms the best non-mixed control on all four medical evaluation rows, improving the average from $43.57\%$ to $46.78\%$. The gains are modest but consistent: $+2.12$ on MedQA, $+2.08$ on MedMCQA, $+4.30$ on PubMedQA, and $+2.47$ on the MMLU medical macro-average. 
The good performance on medical reasoning indicates that the gains from {\method} do not rely on mathematical notation, contest-style structure, or math-specific reward engineering.
It supports the broader TD-Grokking claim: when a domain supplies hard root questions with verifiable answers, training-time decomposition can create useful dense learning signal, leading to better learning outcomes.


%% file: sections/analysis.tex
In this section, we conduct an in-depth analysis of how training subproblems influence the model's behavior on root problems.

\begin{wrapfigure}{r}{0.43\textwidth}
\vspace{-0.8em}
\centering
\includegraphics[width=\linewidth]{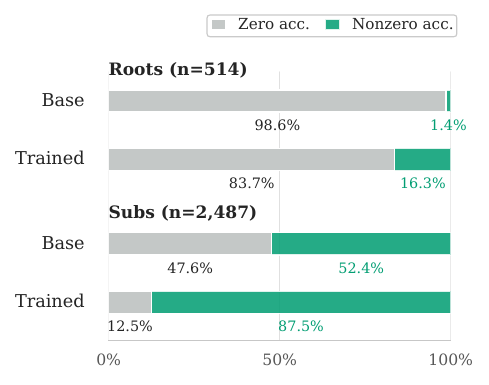}
\caption{
Composition of zero-reward problems before and after sub-only training.
}
\label{fig:zero_acc_composition}
\vspace{-1.0em}
\end{wrapfigure}

\paragraph{Training on subproblems alone can activate learning on root problems.}
As we have already seen in Figure~\ref{fig:mix_root_training_curve} (Left), training only on subproblems can increase the probability that the LLM solves root problems.
Figure~\ref{fig:zero_acc_composition} more clearly shows how zero-reward problems are influenced.
Before training, the base model has zero accuracy on $507$ of $514$ root problems, meaning 98.6\% of roots are in the zero-reward region.
After sub-only RL, $84$ roots have nonzero accuracy. Among these, $77$ are newly recovered from the base zero-reward set, giving a zero-recovery rate of $77/507=15.2\%$.
At the item level, among all the $514$ root questions, $197$ improve, $6$ decline, and $311$ remain unchanged.
Although many zero-reward root problems remain unsolved, sub-only training nevertheless activates root-level learning despite receiving no root-level rewards.



\begin{figure*}[h]
\centering
\includegraphics[width=0.95\textwidth]{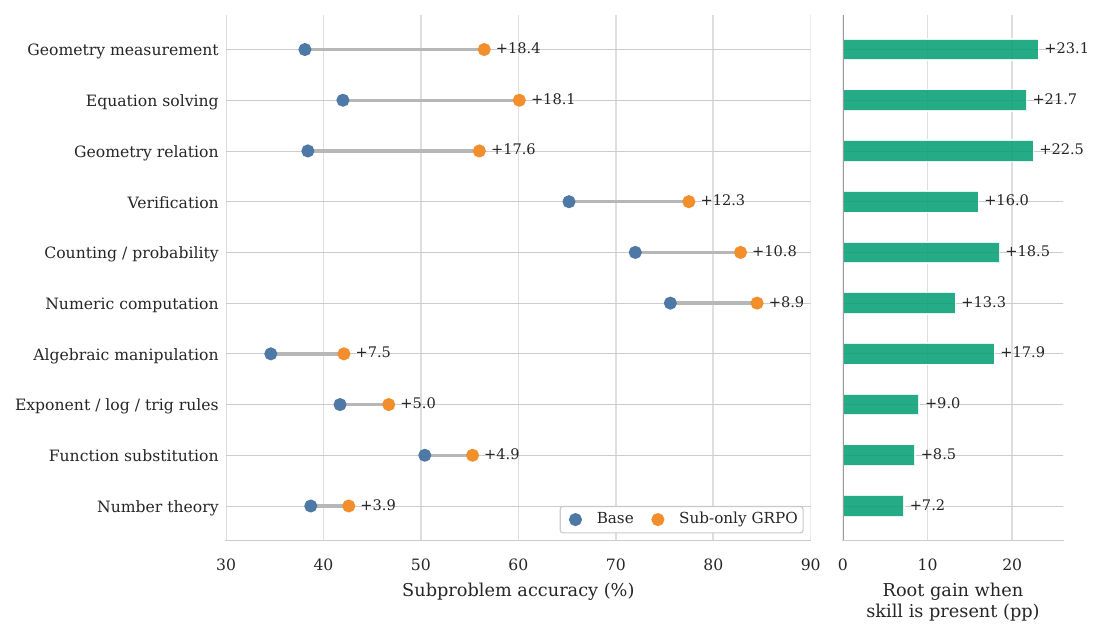}
\caption{
Accuracy changes of subproblems and root problems after sub-only RL in 10 subdomains.
}
\label{fig:base_step190_skill_dumbbell}
\end{figure*}

\paragraph{Higher subproblem gains, higher root problem gains.}
To further verify our hypothesis that learning subskills can directly lead to better solutions to root problems, we investigate changes in model behavior after training from a more fine-grained perspective.
Specifically, we split the DeepScaleR-hard dataset into 10 subdomains to observe the relationship between the performance gains of subproblems and root problems.
As shown in Figure~\ref{fig:base_step190_skill_dumbbell}, the solution rates of both subproblems and root problems increase consistently across all $10$ subdomains, with gains ranging from $3.9$ to $23.1$ percentage points.

\paragraph{Direct training on root problems is still necessary.}
We now check whether successful root-problem solving is merely a deterministic consequence of solving all associated subproblems.
If so, the accuracy of root problems would be a deterministic function of their subproblem accuracies.
Instead, the correlations are positive but weak. Under the trained sub-only checkpoint, Pearson correlations between root accuracy and child-accuracy summaries are $0.25$ for the product, $0.27$ for the minimum, and $0.23$ for the log-product; the corresponding Spearman correlations are $0.28$, $0.30$, and $0.27$.

These weak correlations imply that successful root-problem solving depends not only on mastering subskills, but also on learning more advanced skills, such as choosing the right subskills, planning feasible reasoning paths, and dealing with inconsistencies between different steps.
As a result, mixed training on both subproblems and root problems is necessary to achieve the best performance.




%% file: sections/conclusion.tex
\paragraph{Conclusion.}
We introduced {\method}, a training-time decomposition framework for extracting useful RLVR signal from hard examples that are initially zero-reward under direct outcome supervision.
Instead of changing the verifier, adding process rewards, or providing decomposition at inference time, {\method} converts each hard root into self-contained and verifiable subproblems that can be optimized with the same final-answer reward.
Across mathematical reasoning and medical QA benchmarks, mixed training consistently outperforms direct root-only training, while the sub-only diagnostic shows that root-derived subproblem practice can recover a non-trivial subset of previously unsolved roots even without direct root rewards.
These results support the central claim that decomposition is not only an inference scaffold, but also an effective training-time mechanism for making hard problems learnable.

%% file: sections/limitations.tex
\paragraph{Limitations.}
The main limitation of {\method} is that decomposition is not free: it requires an external construction model, validation passes, and difficulty calibration before RL training begins.
However, our experiments show that this additional computation is worthwhile for zero-reward hard-example pools, where direct root-only training provides little usable signal.
By spending compute up front to expose trainable subproblem tasks, {\method} improves final benchmark performance and the dynamic variant further reduces generated-token cost by retiring mastered subproblems.
Due to limited computational resources, we were not able to conduct broader experiments across more model families, domains, and decomposition quality regimes.

\paragraph{Future work.}
Future work should study recursive decomposition on larger models and harder domains where first-level subproblems may still remain zero-reward.
It would also be useful to develop automatic policies for deciding when the decomposition cost is justified and how much computation should be allocated to construction, validation, calibration, and RL training.

%% file: sections/appendix.tex
\section{Experimental Details}
\label{app:experimental_details}

This appendix records the data construction, benchmark protocol, generation
settings, decomposition prompt, and qualitative cases used in the experiments.
All evaluations are root-prompt evaluations unless a diagnostic is explicitly
described as a subproblem evaluation. The model never receives a decomposition
guide, subproblem list, or intermediate answer at evaluation time.

\subsection{Training Corpus Accounting}
\label{app:corpus}

The mathematical experiments start from the zero-reward slice of
DeepScaleR-hard for Qwen3-1.7B. The initial decomposition pool contains
$4{,}944$ root questions whose root accuracy was zero under the
calibration budget used during data construction. After decomposition,
validation, and salvage, the final training corpus contains $3{,}788$ aligned roots and
$14{,}717$ retained subproblems. Table~\ref{tab:appendix_corpus} gives
the exact training views used by the main mathematical runs.

\begin{table}[h]
\centering
\small
\setlength{\tabcolsep}{5pt}
\begin{tabular}{>{\raggedright\arraybackslash}p{0.22\linewidth}>{\raggedright\arraybackslash}p{0.18\linewidth}>{\raggedright\arraybackslash}p{0.50\linewidth}}
\toprule
\textbf{Training view} & \textbf{Rows} & \textbf{Definition} \\
\midrule
Root-only & $3{,}788$ & Original zero-reward root prompts. Each row keeps the original question and final answer. \\
Sub-only & $14{,}717$ & Flattened self-contained subproblems derived from the same roots. Each subproblem has its own verifiable target answer. \\
Mixed & $18{,}505$ & Union of the vanilla and sub-only rows, preserving the root-subproblem alignment through \texttt{root\_problem\_id}. \\
\bottomrule
\end{tabular}
\caption{Mathematical training corpus used by the reported H100 GRPO runs. The
mixed view has an average of $3.89$ retained subproblems per root.}
\label{tab:appendix_corpus}
\end{table}

The earlier full decomposition pass produced $3{,}989$ usable roots before the
solq95 filter. The paper reports the filtered corpus because it is the dataset
actually used by the H100 root-only, sub-only, mixed, and dynamic runs. The
relationship between the three views is controlled: every retained subproblem is
tied to exactly one retained root, and every retained root appears once in
root-only and mixed training.

\subsection{Formal RLVR Signal View}
\label{app:formal_signal}

This section makes explicit the sparse-reward mechanism that motivates
training-time decomposition. Let a verifiable training item be
$z=(q_z,a_z)$, where $q_z$ is the prompt and $a_z$ is the reference answer.
For rollout $o_{z,k}\sim \pi_\theta(\cdot \mid q_z)$, let
$E(o_{z,k})$ be the extracted final answer and let $V$ be the verifier. The
binary RLVR reward is
\[
r_{z,k}
= \mathbf{1}\!\left[V\!\left(E(o_{z,k}),a_z\right)=1\right].
\]
With $K$ sampled rollouts, the empirical item accuracy is
\[
\widehat{p}_K(z;\theta)
=\frac{1}{K}\sum_{k=1}^{K} r_{z,k}.
\]
A root $x$ is zero-reward for the starting policy if
$\widehat{p}_K(x;\theta_0)=0$. This is not a statement that $x$ is impossible;
it means that the sampled outcome rewards for $x$ contain no positive event
under the current policy and sampling budget.

For GRPO-style group-relative optimization, the outcome advantage for a rollout
from item $z$ is computed from rewards within the same prompt group:
\[
\bar r_z=\frac{1}{G}\sum_{g=1}^{G} r_{z,g},\qquad
s_z=\sqrt{\frac{1}{G}\sum_{g=1}^{G}(r_{z,g}-\bar r_z)^2},
\]
\[
A_{z,g}=\frac{r_{z,g}-\bar r_z}{s_z+\epsilon}.
\]
If all rewards in the group are zero, then $\bar r_z=0$ and the group contains
no outcome-level contrast. If an item has true success probability $p_z$, the
probability that a $G$-rollout group contains at least one positive reward is
\[
P_{\mathrm{info}}(z)=1-(1-p_z)^G .
\]
Thus a zero-reward root with $p_x\approx 0$ is unlikely to produce informative
groups, while a derived subproblem $s$ with $p_s>0$ can produce positive
reward events with probability $1-(1-p_s)^G$. Decomposition densifies RLVR by
moving training mass from roots with $P_{\mathrm{info}}\approx 0$ to
root-conditioned nodes with nonzero $P_{\mathrm{info}}$, while mixed training
keeps some pressure on the original root distribution.

Formally, for each retained root $x$ we construct a one-level decomposition
tree
\[
T_x=\{x\}\cup \mathcal{D}(x),\qquad
\mathcal{D}(x)=\{s_{x,1},\ldots,s_{x,m_x}\}.
\]
Each node $z\in T_x$ has its own verifiable answer $a_z$. The three static
training views are
\[
\mathcal{S}_{\mathrm{root}}=\{x\},\qquad
\mathcal{S}_{\mathrm{sub}}=\bigcup_x \mathcal{D}(x),\qquad
\mathcal{S}_{\mathrm{mix}}=\mathcal{S}_{\mathrm{root}}\cup
\mathcal{S}_{\mathrm{sub}}.
\]
The dynamic variant maintains an active set $\mathcal{A}_t\subseteq
\mathcal{S}_{\mathrm{mix}}$. In the reported implementation,
\[
\widehat{p}_8(z;\theta_t)=0 \Rightarrow
\text{activate children of }z,\qquad
\widehat{p}_8(z;\theta_t)\geq \frac{7}{8} \Rightarrow
\text{retire }z.
\]
These discrete thresholds instantiate the intended ``below 10\%'' and
``above 80\%'' active-set rules under an eight-rollout group.

Finally, the local-to-global intuition can be expressed through a simple
diagnostic approximation. If a root solution requires $m$ local requirements
and requirement $j$ succeeds with probability $p_j$, then under an independence
approximation,
\[
p_{\mathrm{root}}\approx \prod_{j=1}^{m} p_j,\qquad
\Delta \log p_{\mathrm{root}}\approx \sum_{j=1}^{m}
\Delta \log p_j .
\]
This approximation is not an assumption used by the training algorithm; it is a
useful interpretation of the case studies. Improving a weak local step can have
a disproportionate effect on the probability that the full root reasoning
chain closes successfully.

\subsection{Decomposition Pipeline}
\label{app:decomposition_pipeline}

Decomposition is a data-construction step, not an inference-time scaffold. For
each selected root problem, the pipeline obtains or constructs an
answer-consistent guide, segments the guide into local requirements, rewrites
those requirements into standalone problem-answer pairs, and validates the
resulting candidates. Retained candidates must be self-contained and verifiable
by the same final-answer interface as the root problem. The validator also
screens out candidates that simply copy the full root problem or expose the full
solution trace.

The production decomposition generator was DeepSeek-V3.2 through a chat API.
The retry-focused production prompt version was
\texttt{v6\_text\_blocks\_retry\_focus}. The important generation and validation
settings are summarized in Table~\ref{tab:appendix_decomp_params}.

\begin{table}[h]
\centering
\small
\setlength{\tabcolsep}{5pt}
\begin{tabular}{>{\raggedright\arraybackslash}p{0.28\linewidth}>{\raggedright\arraybackslash}p{0.56\linewidth}}
\toprule
\textbf{Component} & \textbf{Setting} \\
\midrule
Generator & DeepSeek-V3.2 via chat-completion API. \\
Prompt style & Natural text-block output; no JSON in the final production prompt. \\
Few-shot selection & One relevant few-shot example in the final retry pass; examples are selected by coarse problem tags and style. \\
Output length & Up to $3{,}000$ tokens in the final retry pass; earlier production passes used smaller limits. \\
Sampling & Temperature $0.2$, top-$p$ $0.95$ for generation of decomposition candidates. \\
Candidate count & Target $4$-$6$ subproblems; allowed range $3$-$8$ unless salvage preserves a longer but usable decomposition. \\
Validation filters & Empty fields, cross-reference phrases, non-self-contained prompts, proof-only prompts, final-answer mismatch, overly verbose answers, duplicate problem-answer pairs, and excessive subproblem counts. \\
\bottomrule
\end{tabular}
\caption{Decomposition generation and validation settings used to build the
mathematical training corpus.}
\label{tab:appendix_decomp_params}
\end{table}

\paragraph{Core decomposition prompt.}
Following the appendix style of full prompt templates, we include the exact
static text of the production decomposition prompt. Problem-dependent slots are
shown in angle brackets. The selected good and bad examples are inserted
verbatim from the few-shot library according to the coarse tags inferred from
the root problem. The \texttt{Verification} field is not used as a reward
signal; it is a construction-time sanity check explaining why the subproblem is
self-contained.

\begin{center}
\textbf{Prompt Template of TD-Grokking Decomposition Problems}
\end{center}

\begin{lstlisting}[style=prompttemplate,caption={System message for decomposition generation.},label={lst:decomp_system_prompt}]
You are a careful math curriculum designer.

Your job is to decompose one original math problem into a small set of useful training subproblems.

You must optimize for:
1. self-contained subproblems,
2. solvable subproblems,
3. independence between subproblems,
4. reusable mathematical skills,
5. exact-answer compatibility with programmatic evaluation.

Output only the requested text blocks.
If a later subproblem needs a quantity that could have been computed earlier, restate that concrete quantity directly inside the new question instead of referring to an earlier subproblem.
\end{lstlisting}

\begin{lstlisting}[style=prompttemplate,caption={User prompt template for decomposition generation.},label={lst:decomp_user_prompt}]
Decompose the following math problem into a small set of training subproblems.

Goal:
- produce subproblems that are independently solvable,
- each subproblem must restate all conditions it needs,
- each subproblem must train a reusable math skill from the original problem,
- the final subproblem must have the same final answer as the original answer.

Hard requirements:
1. Return between 3 and 8 subproblems.
2. Every subproblem must be self-contained. Do not refer to other subproblems, previous results, or hidden context.
3. Every subproblem must restate the numeric values, symbols, domains, constraints, and definitions it needs. This restatement can be natural prose or an explicit `Given:` clause. If a later step reuses a quantity, restate the concrete value or formula directly instead of citing an earlier step.
4. Do not write proof, verification, or "show that" style subproblems.
5. Do not create trivial meta subproblems such as "is the final answer correct?" or "choose the correct option" unless the original problem is inherently multiple-choice and that step still requires real reasoning.
6. Prefer concrete subproblems over placeholder-only algebra. Do not invent abstract variables like S, T, or R unless they are defined inside the same subproblem and genuinely useful.
7. Each `answer` must be a short exact final answer usable as ground truth. Do not include explanations, equations, or multiple sentences in `answer`.
8. Keep `reasoning`, `solution`, and `verification` concise but complete. Never leave any field blank. `verification` must explain why the subproblem is self-contained and solvable on its own, not why it matches another subproblem.
9. The last subproblem must solve for the original target and must end with the original answer exactly.
10. Across the whole decomposition, prefer at least two genuinely different skill steps. Avoid repeating the same shell sentence with only numbers changed.
11. Never return only 1 or 2 subproblems. If the problem feels simple, split it into smaller concrete computations anyway.
12. Do not introduce new helper objects like new polynomials, functions, sequences, points, or variables unless the original problem already uses them or the helper is strictly necessary for a short computation.
13. Do not let the first subproblem absorb the whole task. Each subproblem should ask for one concrete intermediate quantity, relation, or check.
14. Keep each field short, but prioritize completeness over rigid sentence counts: Question at most 3 sentences, Reasoning 1 to 2 short sentences, Solution 1 to 3 short sentences, Answer one line, Verification 1 to 2 short sentences.
15. If unsure, prefer simple arithmetic, algebraic, geometric, or probabilistic intermediate quantities over long theory summaries.
16. If a subproblem would otherwise be too long, shorten the decomposition or simplify the wording. Do not leave `solution`, `answer`, or `verification` blank.
17. Each subproblem should target exactly one concrete intermediate quantity, one concrete relation, or one concrete counting/probability result. Do not turn a subproblem into a mini-lecture or a long multi-part derivation.
18. Target 4 to 6 subproblems by default. Use 7 or 8 only when the original problem clearly needs them. Never exceed 8.
19. If a later subproblem needs an earlier result, rewrite the question in the form `Given <explicit value or formula> ...` or restate the full quantity in plain language. Never write phrases like `from the previous step`, `from Subproblem 3`, `using the result above`, or `same as before`.
20. If you are running out of space, reduce the number of subproblems instead of leaving the last subproblem incomplete. The final subproblem must always contain a non-empty `solution` and a non-empty `answer`.
21. The final `answer` should match the original answer's target and outer form as closely as possible. If the original answer is an expression, give only that expression. If it is an equation, inequality, set, ordered pair, list, or named quantity, preserve that outer structure instead of answering with a different but related object.
22. Do not end with a verification-only subproblem. If the last step would only check correctness, merge it into the previous computational step and keep the final answer there.
23. In `answer`, write only the target answer. Do not write explanatory prefixes such as `Therefore`, `So`, `The answer is`, `check`, `because`, or a full sentence.

Style preference:
<STYLE_INSTRUCTION>

Return only text in this exact structure:

### Subproblem 1
Question: ...
Reasoning: ...
Solution: ...
Answer: ...
Verification: ...

### Subproblem 2
Question: ...
Reasoning: ...
Solution: ...
Answer: ...
Verification: ...

### Subproblem 3
Question: ...
Reasoning: ...
Solution: ...
Answer: ...
Verification: ...

Use the same six labels for every subproblem block. Do not use JSON. Do not use code fences.

Good decomposition examples to imitate in spirit, not in wording:
<SELECTED_GOOD_DECOMPOSITION_EXAMPLES>

Bad decomposition pattern to avoid:
<SELECTED_BAD_DECOMPOSITION_EXAMPLES>

Original problem:
<ORIGINAL_PROBLEM>

Original answer:
<ORIGINAL_ANSWER>

Reference solution:
<REFERENCE_SOLUTION_OR_NO_REFERENCE_SOLUTION_PROVIDED>

The End of Prompt
\end{lstlisting}

For the final production retry pass, \texttt{STYLE\_INSTRUCTION} was the
natural-style instruction:
\begin{quote}
\small
Prefer natural question wording. Restate necessary conditions inside each
question, but do not force every subproblem into the same explicit template.
Use \texttt{Given:} only when it genuinely improves clarity.
\end{quote}

\subsection{Subproblem Skill Annotation Prompt}
\label{app:skill_annotation_prompt}

The skill labels used in the diagnostic figures are produced after
decomposition. The annotation pass is separate from training and evaluation:
the labels are used only for analysis, not as policy input, reward input, or
test-time scaffolding. We use a frozen taxonomy with one primary label and up
to two secondary labels per subproblem. The exact prompt template used for the
annotation pass is shown below.

\begin{lstlisting}[style=prompttemplate,caption={System message for atomic skill annotation.},label={lst:skill_classification_system_prompt}]
You are a careful math annotation assistant.

Your task is to classify each subproblem into an atomic skill taxonomy.

You must follow these rules strictly:

1. Use only labels from the provided label set.
2. Output exactly one primary_label.
3. Output zero to two secondary_labels.
4. Treat verification as a secondary label unless the problem is almost entirely a consistency check.
5. Prefer the skill required to solve the problem, not superficial wording.
6. If the problem asks for an unknown value by solving an equation or system, prefer equation_solving.
7. If the problem mainly asks to rewrite, simplify, or transform an expression, prefer algebraic_manipulation.
8. If the problem is mainly about ratios, averages, unit rates, prices, speed, or percentages, prefer ratio_rate_percent.
9. If the problem is mainly about counting, probability, expectation, or combinatorial arrangements, prefer counting_probability.
10. If the problem is mainly about divisibility, modular arithmetic, bases, remainders, gcd, lcm, or integer structure, prefer number_theory.
11. If the problem is mainly about geometric quantities such as length, area, volume, perimeter, coordinates, or chord length, prefer geometry_measurement.
12. If the problem is mainly about geometric theorems, angle relations, tangent properties, conics, similarity, congruence, or geometric structure, prefer geometry_relation.
13. If the problem is mainly direct evaluation of a known numeric expression, prefer numeric_computation.
14. If the problem is mainly about recurrence, progression, repeated updates, or iterative processes, prefer sequence_recurrence.
15. If the problem is mainly about function definition recovery or substitution into a function form, prefer function_substitution.
16. If the problem is mainly about exponent rules, logarithm rules, trigonometric identities, or standard trigonometric ratios/values, prefer exponent_log_trig_rules.

Output valid JSON only.

Return:
{
  "results": [
    {
      "problem_id": "...",
      "primary_label": "...",
      "secondary_labels": ["...", "..."],
      "rationale": "..."
    }
  ]
}
\end{lstlisting}

\begin{lstlisting}[style=prompttemplate,caption={User prompt template for atomic skill annotation.},label={lst:skill_classification_user_prompt}]
Classify the following subproblems using the fixed atomic skill taxonomy.

Allowed labels:
- numeric_computation
- algebraic_manipulation
- equation_solving
- function_substitution
- exponent_log_trig_rules
- ratio_rate_percent
- counting_probability
- number_theory
- sequence_recurrence
- geometry_measurement
- geometry_relation
- verification

Few-shot examples:
<FEWSHOT_JSON_LINES>

Target batch:
{
  "label_set": [
    "numeric_computation",
    "algebraic_manipulation",
    "equation_solving",
    "function_substitution",
    "exponent_log_trig_rules",
    "ratio_rate_percent",
    "counting_probability",
    "number_theory",
    "sequence_recurrence",
    "geometry_measurement",
    "geometry_relation",
    "verification"
  ],
  "subproblems": [
    {
      "problem_id": "...",
      "question": "..."
    }
  ]
}
\end{lstlisting}

\begin{figure*}[t]
\centering
\includegraphics[width=0.86\textwidth]{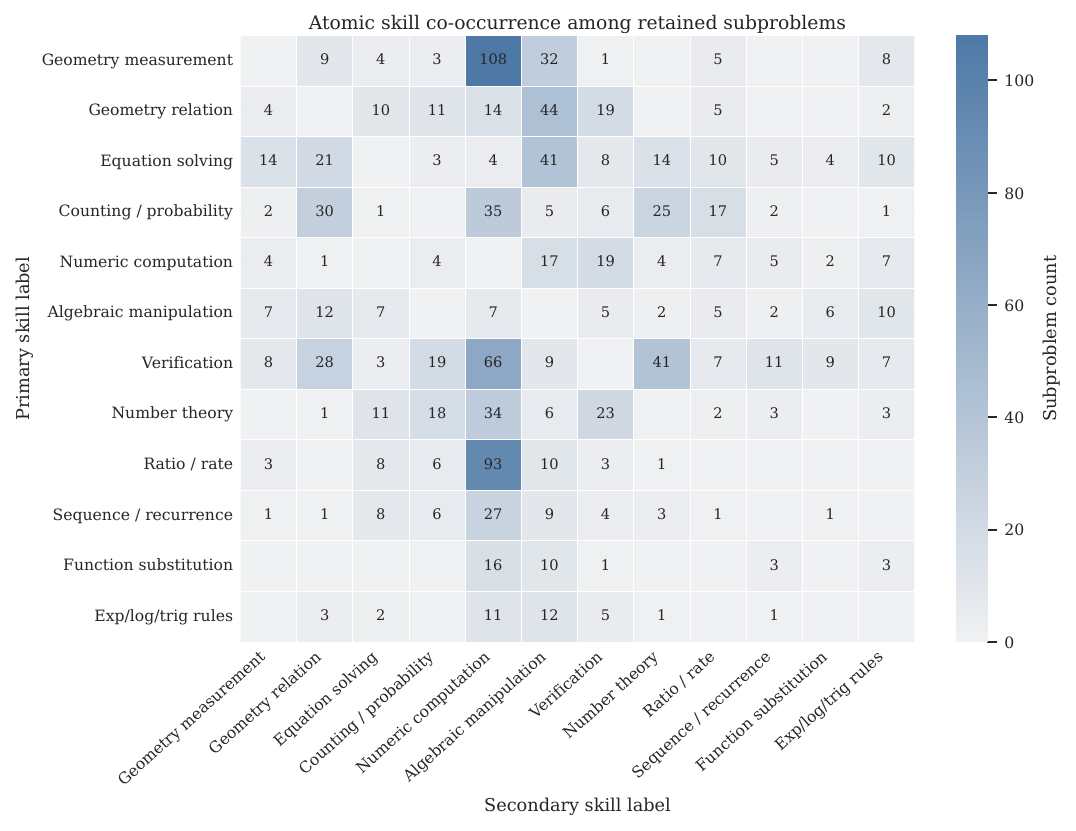}
\caption{
Primary-secondary atomic skill co-occurrence over the $2{,}487$ retained
subproblems in the paired diagnostic. Rows are primary labels and columns are
secondary labels; blank cells have zero count. The same frozen taxonomy is used
for the skill-conditioned transfer analysis in
Figure~\ref{fig:base_step190_skill_dumbbell} and
Table~\ref{tab:skill_transfer}.
}
\label{fig:appendix_skill_label_heatmap}
\end{figure*}

The final retained diagnostic set contains $2{,}487$ labeled subproblems from
$514$ parent roots. Figure~\ref{fig:appendix_skill_label_heatmap} visualizes
the primary-secondary co-occurrence structure induced by the annotation prompt.
The heatmap is sparse by design: secondary labels are used only when a
subproblem genuinely requires a second skill rather than as a generic topic
tag.

\clearpage

\subsection{Dynamic Active-Set Rule}
\label{app:dynamic_rule}

The dynamic variant uses the same one-level root-subproblem graph as the mixed
training view, but changes which rows are active. Each root starts active. A
root with no successful rollout in the current $8$-sample group activates its
first-level subproblems. A row with at least $7$ successful rollouts out of $8$
is treated as mastered and can leave the active set. Thus the textual thresholds
``below 10\%'' and ``above 80\%'' are implemented as the discrete rules $0/8$
and at least $7/8$, respectively. This keeps the reward function unchanged and
only changes the training sampler.

\subsection{Benchmark and Generation Protocols}
\label{app:benchmarks}

All mathematical benchmark results in Table~\ref{tab:main_math_results} are
produced through the same local lm-eval/vLLM evaluation wrapper. Unless stated
otherwise, generation uses Qwen thinking mode with chat templating enabled,
temperature $0.6$, top-$p$ $0.95$, top-$k$ $20$, min-$p$ $0$, and bfloat16
vLLM inference. AIME results use repeated sampling; the other mathematical
benchmarks report single-run exact-match accuracy.

\begin{table*}[t]
\caption{Benchmark protocols used for the mathematical evaluation suite.}
\label{tab:appendix_math_benchmarks}
\centering
\small
\setlength{\tabcolsep}{3.5pt}
\resizebox{\textwidth}{!}{%
\begin{tabular}{>{\raggedright\arraybackslash}p{0.16\linewidth}>{\raggedright\arraybackslash}p{0.11\linewidth}>{\raggedright\arraybackslash}p{0.19\linewidth}>{\raggedright\arraybackslash}p{0.35\linewidth}}
\toprule
\textbf{Benchmark} & \textbf{Items} & \textbf{Metric} & \textbf{Generation and scoring notes} \\
\midrule
AIME 2024 & $30$ & avg@8, pass@8 & Eight sampled completions per problem. Main runs use max generation $38{,}912$ tokens and exact-answer extraction from the final boxed or normalized answer. \\
AIME 2025 & $30$ & avg@8, pass@8 & Same repeated-sampling protocol as AIME 2024. pass@8 marks a problem correct if any of the eight completions verifies. \\
MATH500 & $500$ & pass@1 & Single completion per item. Core runs used the same thinking-mode wrapper; long-generation runs used a $32{,}768$ to $38{,}912$ token cap depending on the run family. \\
OlympiadBench & $674$ & pass@1 & Math subset through the local external-math task. Scored with the same final-answer extraction and equivalence checker as the root verifier. \\
AMC23 & $40$ & avg@8 & Local external-math task over the AMC23 split; official-aligned supplement runs use max generation $38{,}912$ tokens. \\
\bottomrule
\end{tabular}%
}
\end{table*}

For the medical-domain instantiation, we use generative local tasks that ask the
model to produce the final option or short answer directly. The medical
evaluation does not use the math boxed-answer system instruction. Reported
medical scores are exact-match percentages after task-specific normalization.
The MMLU Medical row in Table~\ref{tab:medical_domain} is the macro-average of
Anatomy, Clinical Knowledge, College Medicine, Medical Genetics, and
Professional Medicine.

\begin{table}[h]
\centering
\small
\setlength{\tabcolsep}{4pt}
\begin{tabular}{>{\raggedright\arraybackslash}p{0.35\linewidth}>{\raggedright\arraybackslash}p{0.17\linewidth}>{\raggedright\arraybackslash}p{0.36\linewidth}}
\toprule
\textbf{Medical benchmark} & \textbf{Items} & \textbf{Scoring} \\
\midrule
MedQA-USMLE & $1{,}273$ & Exact match on the generated option or answer. \\
MedMCQA & $4{,}183$ & Exact match on the generated option or answer. \\
PubMedQA & $1{,}000$ & Exact match over yes/no/maybe-style generated answers. \\
MMLU Medical Avg. & Five subsets & Macro-average over five generated MMLU medical tasks. \\
\bottomrule
\end{tabular}
\caption{Medical-domain benchmark protocol.}
\label{tab:appendix_med_benchmarks}
\end{table}

\subsection{Training Hyperparameters}
\label{app:training_hparams}

The H100 mathematical GRPO runs share the same training skeleton wherever
possible. The base model is Qwen3-1.7B, the reward is the existing final-answer
math verifier, and no decomposition-specific reward shaping is introduced.
Unless otherwise noted, the reported mathematical RL experiments were run on a
single server with $8$ NVIDIA H100 GPUs.
Table~\ref{tab:appendix_training_hparams} lists the stable run parameters used
by the root-only, sub-only, mixed, and dynamic families.

\begin{table}[h]
\centering
\small
\setlength{\tabcolsep}{5pt}
\begin{tabular}{>{\raggedright\arraybackslash}p{0.30\linewidth}>{\raggedright\arraybackslash}p{0.52\linewidth}}
\toprule
\textbf{Parameter} & \textbf{Value} \\
\midrule
Base checkpoint & Qwen3-1.7B, non-Base instruct model. \\
RL algorithm & GRPO with group-relative outcome advantages. \\
Reward & Binary final-answer verification through the existing math reward path. \\
GPUs & One $8$-GPU H100 node for the reported main mathematical runs. \\
Train batch size & $64$. \\
Rollouts per prompt & $8$ for dynamic and later H100 runs; earlier static runs used the same GRPO family unless otherwise noted. \\
Reward workers & $64$ CPU reward workers in the H100 launchers. \\
Maximum response length & $16{,}384$ tokens during training rollouts. \\
Sampling during training & Temperature $1.0$, top-$p$ $0.95$, top-$k=-1$ for the GRPO rollout generator in the mixed solq95 family. \\
\bottomrule
\end{tabular}
\caption{Main mathematical GRPO training settings. The evaluation sampling
parameters differ from training and are reported separately in
Appendix~\ref{app:benchmarks}.}
\label{tab:appendix_training_hparams}
\end{table}

\subsection{Anonymized Artifact}
\label{app:artifact}

The anonymized code and data artifact is available at
\url{https://anonymous.4open.science/r/TD-Grokking-6567}. It includes the
data-construction prompts, preprocessing scripts, training launchers,
evaluation scripts, and configuration files needed to reproduce the main
experimental comparisons.

\section{Subproblem Skill Diagnostic}
\label{app:skill_diagnostic}

Before the full H100 benchmark suite, we ran a paired diagnostic to test whether
training only on subproblems can improve both subproblem accuracy and the
corresponding root accuracy. The diagnostic contains $514$ root problems and
$2{,}487$ subproblems. Each item is evaluated with $64$ rollouts,
temperature $0.6$, top-$p$ $0.95$, and a $16$k token cap. In this diagnostic,
strict exact-match accuracy rises from $1.4\%$ to $16.3\%$ on roots and from
$52.3\%$ to $75.2\%$ on subproblems.
Table~\ref{tab:skill_transfer} reports the full skill-conditioned transfer
breakdown.

\begin{table}[h]
\centering
\small
\setlength{\tabcolsep}{4pt}
\begin{tabular}{lrrrr}
\toprule
\textbf{Skill} & \textbf{\#Roots} & \textbf{Subproblem $\Delta$} & \textbf{Root $\Delta$} & \textbf{Gap} \\
\midrule
Geometry measurement & 109 & +18.4 & +23.1 & +10.1 \\
Equation solving & 134 & +18.1 & +21.7 & +8.2 \\
Geometry relation & 83 & +17.6 & +22.5 & +8.8 \\
Verification & 190 & +12.3 & +16.0 & +1.0 \\
Counting / probability & 137 & +10.8 & +18.5 & +4.3 \\
Numeric computation & 241 & +8.9 & +13.3 & -2.2 \\
Algebraic manipulation & 97 & +7.5 & +17.9 & +3.6 \\
Exponent / log / trig rules & 25 & +5.0 & +9.0 & -8.0 \\
Function substitution & 20 & +4.9 & +8.5 & -8.5 \\
Number theory & 80 & +3.9 & +7.2 & -9.2 \\
\bottomrule
\end{tabular}
\caption{
Skill-conditioned transfer after sub-only RL. All deltas are percentage points under strict exact-match evaluation. Subproblem $\Delta$ uses the same subproblem skill rows as Figure~\ref{fig:base_step190_skill_dumbbell}. Root $\Delta$ is the mean root-accuracy improvement for roots containing the skill; Gap compares that root delta with roots not containing the skill. Skill groups overlap, so these conditional gains are not additive.
}
\label{tab:skill_transfer}
\end{table}

\begin{figure*}[t]
\centering
\includegraphics[width=0.86\textwidth]{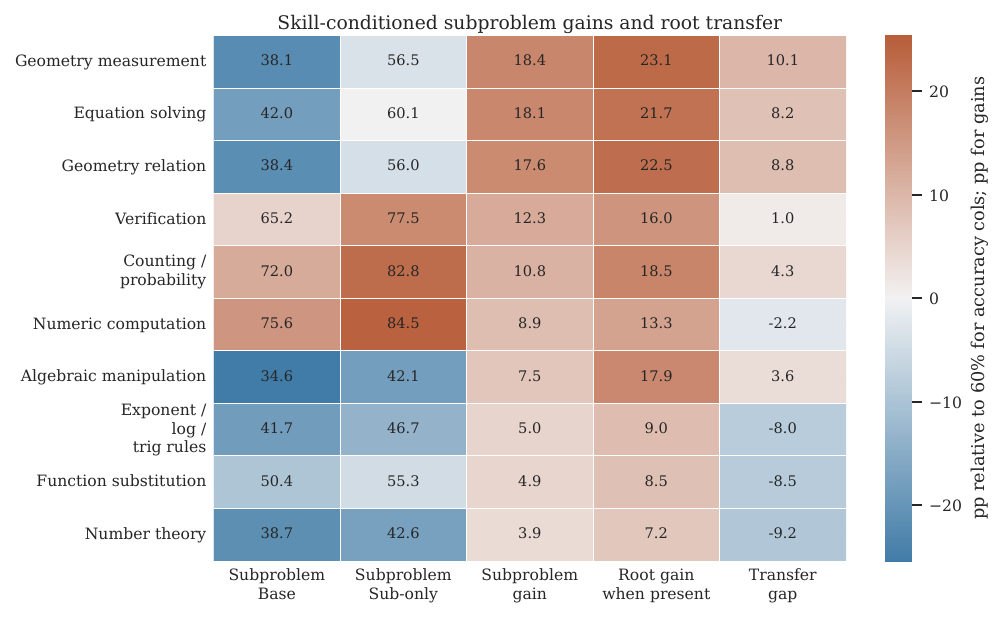}
\caption{
Heatmap view of the skill-conditioned transfer table. The first two columns
show strict subproblem accuracy for the base model and the Sub-only GRPO
checkpoint; the remaining columns show subproblem gain, root gain for roots
containing the skill, and the conditional transfer gap. The values are the same
as Table~\ref{tab:skill_transfer} and the source CSV is included with the
figure artifacts.
}
\label{fig:appendix_skill_transfer_heatmap}
\end{figure*}

\begin{figure}[h]
\centering
\includegraphics[width=0.72\linewidth]{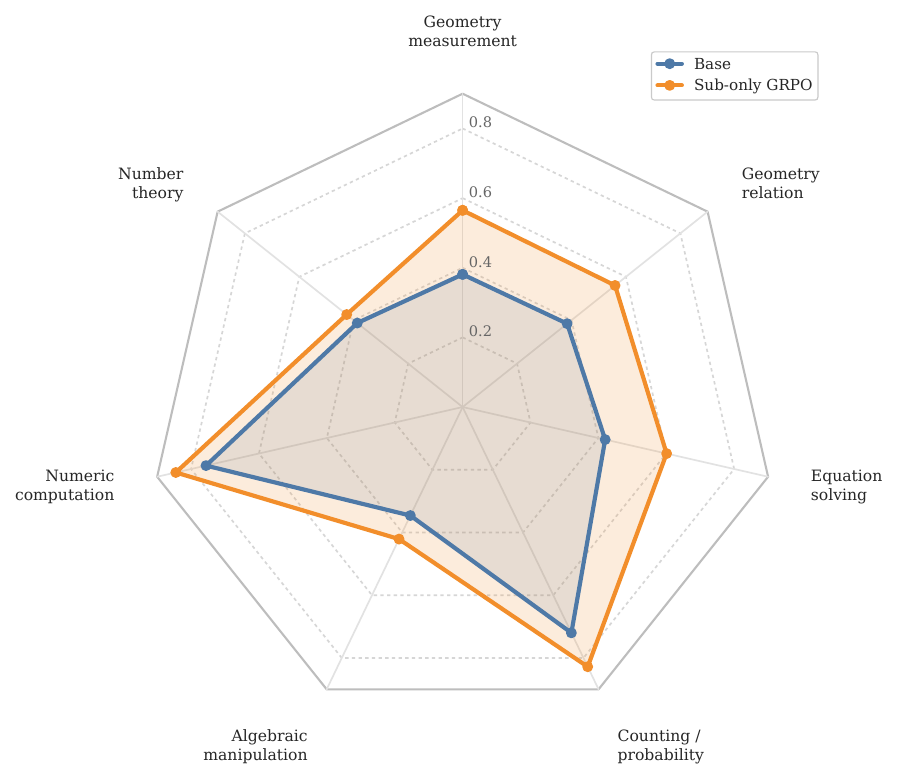}
\caption{Subproblem skill radar for the paired diagnostic. The subproblem-only
checkpoint improves strict accuracy across the seven core subproblem skill
families shown here, with especially large gains on geometry, equation-solving,
counting/probability, and numeric computation.}
\label{fig:appendix_skill_radar}
\end{figure}

The radar is intended as a compact visual summary, while the main quantitative
claim remains root-prompt performance. Several skill-level patterns are useful
for interpretation. In the strict exact-match transfer table,
geometry-measurement roots improve by $+23.1$ percentage points when that skill
is present; geometry-relation roots improve by $+22.5$ points;
equation-solving roots improve by $+21.7$ points; and counting/probability
roots improve by $+18.5$ points. By contrast, number-theory roots improve less
and have a negative transfer gap even though their subproblem scores improve.
This is consistent with the paper's claim that subproblem training helps by
stabilizing local steps, but does not guarantee uniform root-level transfer for
every skill family.

This asymmetry argues against a simple ``subproblem accuracy goes up, therefore
root accuracy goes up'' explanation. Transfer is strongest when subproblems
match local bottlenecks, while number-theory roots remain difficult, likely
because they require global organization or insights not captured by the current
subproblems.

\section{Case Studies}
\label{app:case_studies}

We selected case studies from a held-out diagnostic subset with saved rollout
text. The goal is not to show isolated anecdotes as the main evidence, but to
make the mechanism behind the aggregate gains inspectable. The favorable cases
in Table~\ref{tab:appendix_case_summary} share the same pattern: after
subproblem-only training, the model more often commits to a viable solution
path, keeps key intermediate values stable, recovers from local mistakes, and
closes the reasoning chain to the final answer.

\begin{table*}[t]
\caption{Representative positive-transfer cases from the saved rollout-text
subset. Scores are pass@1 estimates over $64$ sampled rollouts.}
\label{tab:appendix_case_summary}
\centering
\small
\setlength{\tabcolsep}{3.5pt}
\resizebox{\textwidth}{!}{%
\begin{tabular}{>{\raggedright\arraybackslash}p{0.15\linewidth}>{\raggedright\arraybackslash}p{0.19\linewidth}ccc >{\raggedright\arraybackslash}p{0.39\linewidth}}
\toprule
\textbf{Root id} & \textbf{Skill family} & \textbf{Base} & \textbf{Sub-only GRPO} & \textbf{Gain} & \textbf{Mechanism visible in rollouts} \\
\midrule
\texttt{00000900} & Geometry measurement + relation & $20.3$ & $79.7$ & $+59.4$ & The trained model more consistently chooses a coordinate solution for the 13-14-15 triangle, computes the midpoint/circle intersection, and closes all three distances to $195/8$. \\
\texttt{00000631} & Counting/probability + ratio & $43.8$ & $76.6$ & $+32.8$ & The trained model more often corrects the common sample-space error and returns to the $3 \times 3=9$ left-right shoe-pair space, yielding probability $1/3$. \\
\texttt{00000312} & Counting/probability + ratio & $15.6$ & $51.6$ & $+35.9$ & The trained model stabilizes the dice-sum count $2052$ and simplifies $2052/4320$ to $19/40$ rather than oscillating between inconsistent intermediate totals. \\
\texttt{00000454} & Counting/probability + numeric computation & $0.0$ & $34.4$ & $+34.4$ & The trained model learns to condition on fixed group membership, choose the remaining dogs in the constrained groups, and multiply the binomial factors. \\
\texttt{00000625} & Geometry measurement + relation & $4.7$ & $57.8$ & $+53.1$ & The trained model more reliably extracts the relevant coordinates from the diagram and computes $AG$ with the distance formula. \\
\bottomrule
\end{tabular}%
}
\end{table*}

\begin{figure}[h]
\centering
\includegraphics[width=0.78\linewidth]{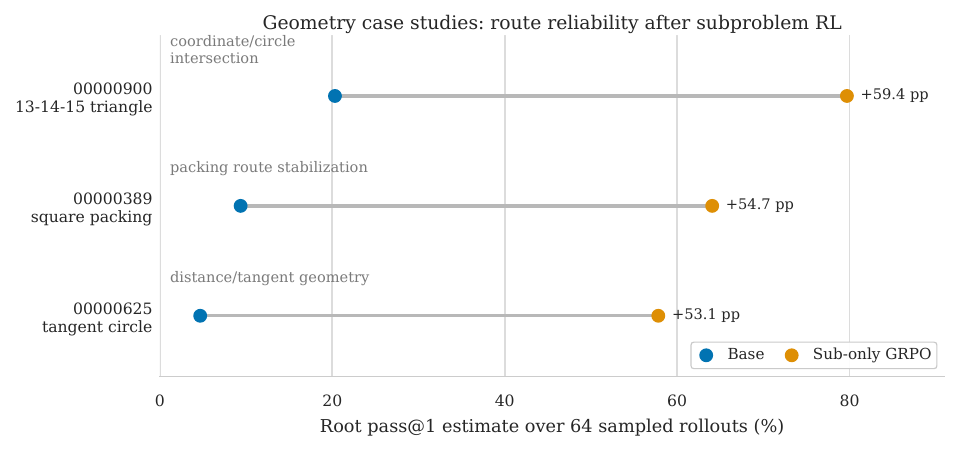}
\caption{
Geometry case-study gains under subproblem-only RL. Each row is a root problem
with saved rollout text; points show the base and Sub-only GRPO root pass@1
estimates over $64$ sampled rollouts. These are the geometry rows from
Table~\ref{tab:appendix_case_summary}, emphasizing that local geometry practice
can make a previously unreliable full-solution route much more repeatable.
}
\label{fig:appendix_geometry_case_study_panel}
\end{figure}

\paragraph{Geometry case.}
For \texttt{deepscaler\_00000900}, the base model is not uniformly incapable:
some rollouts find the correct route. The failure mode is instability: the
model alternates between coordinate geometry, special-point guesses, and
unfinished circle reasoning. After subproblem-only training, the successful
route appears much more often: place the triangle at coordinates, compute
midpoints, derive the two circumcircles, find the second intersection, and use
the distance formula. This supports the interpretation that subproblem practice
does not merely teach a final answer format; it makes a multi-step route more
repeatable.

\paragraph{Probability cases.}
For \texttt{deepscaler\_00000631}, the main error is a wrong sample space. The
trained model is more likely to notice the inconsistency and return to the
three-left-shoes by three-right-shoes sample space. For
\texttt{deepscaler\_00000312}, the key improvement is intermediate-count
stability: base rollouts often start with the right summation but later disagree
with themselves about the favorable count, while the trained model more often
keeps the count $2052$ through the final simplification.

\paragraph{Boundary of the case evidence.}
The same saved case-study set also contains apparent negative cases, but several
of them are contaminated by equivalent-answer or ground-truth issues. For
example, a time-answer case alternates between seconds and minutes-seconds
forms, and two number-theory cases have likely answer conflicts in the stored
ground truth. We therefore use the case-study section to explain positive
transfer mechanisms and avoid claiming a clean taxonomy of negative transfer
from these few text-inspectable examples.

\section{Evaluation and Data Quality Notes}
\label{app:quality_notes}

The main experiments use exact or verifier-based final-answer scoring. This is
appropriate for RLVR but introduces several practical quality issues that we
track explicitly.

\paragraph{Answer extraction.}
For math benchmarks, outputs are reduced to final answers using boxed-answer
extraction plus local normalization and equivalence checks. Benchmark-specific
scorer logs record any blank targets or normalization failures before aggregate
reporting.

\paragraph{Generation mismatch.}
Training rollouts and benchmark evaluations intentionally use different
sampling regimes. Training in the mixed solq95 family uses temperature $1.0$,
top-$p$ $0.95$, top-$k=-1$, and a $16$k response cap. The main benchmark
evaluations use Qwen thinking-mode decoding with temperature $0.6$, top-$p$
$0.95$, top-$k$ $20$, min-$p$ $0$, and longer generation caps. This mismatch is
reported because it affects output length and stability, especially on harder
contest-style benchmarks.

\paragraph{No test-time decomposition.}
All benchmark results are evaluated on the original prompts. Decomposition
guides, subproblem prompts, subproblem answers, and validation metadata are
strictly training-time artifacts.

\paragraph{Existing assets.}
We use Qwen3-1.7B, DeepScaleR, verl, vLLM, lm-eval-harness, and public mathematical reasoning benchmarks including MATH500, AIME, AMC, OlympiadBench, and Omni-MATH. For each existing asset, we cite the original source and report the corresponding version, URL, license, and terms of use where available. We use these assets only for research and evaluation purposes and follow their stated licenses and usage terms.